\pdfoutput=1

\documentclass[table,xcdraw,11pt]{article}

\usepackage{acl}

\usepackage{times}
\usepackage{latexsym}
\usepackage{setspace}
\newcommand\blfootnote[1]{%
  \begingroup
  \renewcommand\thefootnote{}\footnote{#1}%
  \addtocounter{footnote}{-1}%
  \endgroup
}

\usepackage[T1]{fontenc}

\usepackage[utf8]{inputenc}

\usepackage{microtype}

\usepackage{xcolor}

\usepackage{tabularx}
\usepackage{booktabs}
\usepackage{multirow}

\usepackage{stfloats}

\usepackage{listings}
\definecolor{codegreen}{rgb}{0,0.6,0}
\definecolor{codegray}{rgb}{0.5,0.5,0.5}
\definecolor{codepurple}{rgb}{0.58,0,0.82}
\definecolor{backcolour}{HTML}{EEEEEE}

\lstdefinestyle{mystyleA}{
    backgroundcolor=\color{backcolour},   
    commentstyle=\color{codegreen},
    keywordstyle=\color{magenta},
    numberstyle=\tiny\color{codegray},
    stringstyle=\color{codepurple},
    basicstyle=\ttfamily\scriptsize,
    breakatwhitespace=false,         
    breaklines=true,                 
    captionpos=b,                    
    keepspaces=true,                 
    numbers=left,                    
    numbersep=5pt,                  
    showspaces=false,                
    showstringspaces=false,
    showtabs=false,                  
    tabsize=2
}

\colorlet{punct}{red!60!black}
\definecolor{background}{HTML}{EEEEEE}
\definecolor{delim}{RGB}{20,105,176}
\colorlet{numb}{magenta!60!black}

\lstdefinelanguage{json}{
    basicstyle=\ttfamily\scriptsize,
    showstringspaces=false,
    breaklines=true,
    backgroundcolor=\color{background},
    literate=
      {:}{{{\color{punct}{:}}}}{1}
      {,}{{{\color{punct}{,}}}}{1}
      {\{}{{{\color{delim}{\{}}}}{1}
      {\}}{{{\color{delim}{\}}}}}{1}
      {[}{{{\color{delim}{[}}}}{1}
      {]}{{{\color{delim}{]}}}}{1},
}
\usepackage{graphicx}
\usepackage[normalem]{ulem}
\useunder{\uline}{\ul}{}

\title{TimeLMs: Diachronic Language Models from Twitter}

\author{\textbf{Daniel Loureiro*$^{\spadesuit}$, Francesco Barbieri*$^{\clubsuit}$,}\\\textbf{Leonardo Neves$^{\clubsuit}$, Luis Espinosa Anke$^{\diamondsuit}$, Jose Camacho-Collados$^{\diamondsuit}$} \\
$^{\spadesuit}$   LIAAD - INESC TEC, University of Porto, Portugal \\
 $^{\clubsuit}$ Snap Inc., Santa Monica, California, USA  \\ 
  $^{\diamondsuit}$ Cardiff NLP, School of Computer Science and Informatics, Cardiff University, UK  \\
  $^{\spadesuit}$ \texttt{daniel.b.loureiro@inesctec.pt} , $^{\clubsuit}$ \texttt{\{fbarbieri,lneves\}@snap.com}, \\
   $^{\diamondsuit}$ \texttt{\{espinosa-ankel,camachocolladosj\}@cardiff.ac.uk} 
 }

\begin{document}
\maketitle

\begin{abstract}
Despite its importance, the \textit{time} variable has been largely neglected in the NLP and language model literature. In this paper, we present TimeLMs, a set of language models specialized on diachronic Twitter data. We show that a continual learning strategy contributes to enhancing Twitter-based language models' capacity to deal with future and out-of-distribution tweets, while making them competitive with standardized and more monolithic benchmarks. We also perform a number of qualitative analyses showing how they cope with trends and peaks in activity involving specific named entities or concept drift. TimeLMs is available at \url{https://github.com/cardiffnlp/timelms}. \blfootnote{Authors marked with an asterisk (*) contributed equally.}
\end{abstract}


\section{Introduction}

Neural language models (LMs) \cite{devlin-etal-2019-bert,radford2019language, liu2019roberta} are today a key enabler in NLP. They have contributed to a general uplift in downstream performance across many applications, even sometimes rivaling human judgement \cite{wang2018glue,wang2019superglue}, while also bringing about a new paradigm of knowledge acquisition through pre-training. However, currently, both from model development and evaluation standpoints, this paradigm is essentially static, which affects both the ability to generalize to future data and the reliability of experimental results, since it is not uncommon that evaluation benchmarks overlap with pre-training corpora \cite{lazaridou2021pitfalls}. As an example, neither the original versions of BERT and RoBERTa are up to date with the current coronavirus pandemic. This is clearly troublesome, as most of the communication in recent years has been affected by it, yet these models would barely know what we are referring to when we talk about 
\textit{COVID-19} or \textit{lockdown}, to name just a few examples. The lack of \textit{diachronic specialization} is especially concerning in contexts such as social media, where topics of discussion change often and rapidly \cite{del-tredici-etal-2019-short}.


In this paper, we address this issue by sharing with the community a series of \textit{time-specific LMs specialized to Twitter data} (TimeLMs). Our initiative goes beyond the initial release, analysis and experimental results reported in this paper, as models will periodically continue to be trained, improved and released.

\section{Related Work}

There exists a significant body of work on dealing with the time variable in NLP. For instance, by specializing language representations derived from word embedding models or neural networks \cite{hamilton-etal-2016-diachronic,szymanski-2017-temporal,rosenfeld-erk-2018-deep,del-tredici-etal-2019-short,hofmann-etal-2021-dynamic}. Concerning the particular case of LMs, exposing them to new data and updating their parameters accordingly, also known as continual learning, is a promising direction, with an established tradition in machine learning \cite{NIPS2017_f8752278,lewis2020question,lazaridou2021pitfalls,jang2021towards}. Other works, however, have proposed to enhance BERT-based topic models with the time variable \cite{grootendorst2020bertopic}. With regards to in-domain specialization, there are numerous approaches that perform domain adaptation by pre-training a generic LM on specialized corpora. A well-known case is the biomedical domain, e.g., BioBERT \cite{lee2020biobert}, SciBERT \cite{beltagy-etal-2019-scibert} or PubMedBERT \cite{gu2021domain}. In addition to these approaches to specialize language models, there have been similar temporal adaptation analyses to the one presented in our paper \cite{agarwal2021temporal,jin2021lifelong}. In particular, these works showed that training language models in recent data can be beneficial, an improvement that was found to be marginal in \newcite{luu2021time} in a different setting. In terms of continual lifelong learning, which is tangential to our main goal, \newcite{biesialska2020continual} provide a detailed survey on the main techniques proposed in the NLP literature.

More relevant to this paper, on the other hand, are LMs specialized to social media data, specifically Twitter, with BERTweet \cite{nguyen-etal-2020-bertweet}, TweetEval \cite{barbieri-etal-2020-tweeteval} and XLM-T \cite{barbieri2021xlm} being, to the best of our knowledge, the most prominent examples. However, the above efforts barely address the diachronic nature of language. 
Crucially, they do not address the problem of specializing LMs to social media \textit{and} putting the time variable at the core of the framework. Moreover, it is desirable that such time-aware models are released alongside usable software and a reliable infrastructure.
Our TimeLMs initiative, detailed in Section \ref{sec:timelms}, aims to address the above challenges.


\section{TimeLMs: Diachronic Language Models from Twitter}
\label{sec:timelms}

In this section, we present our approach to train language models for different time periods.


\subsection{Twitter corpus}
\label{sec:corpus}

For the training and evaluation of language models, we first collect a large corpus of tweets. 
In the following we explain both the data collection and cleaning processes. 


\noindent \textbf{Data collection.}
We use the Twitter Academic API to obtain a large sample of tweets evenly distributed across time.
In order to obtain a sample which is representative of general conversation on that social platform, we query the API using the most frequent stopwords\footnote{We use the top 10 entries from: \href{https://github.com/first20hours/google-10000-english/raw/master/google-10000-english.txt}{google-10000-english.txt}}, for a set number of tweets at timestamps distanced by 5 minutes - for every hour of every day constituting a particular yearly quarter.
We also use specific flags supported by the API to retrieve only tweets in English and ignore retweets, quotes, links, media posts and ads.

For our initial base model (2019-90M henceforth), we used an evenly time-distributed corpus from the API, for the period between 2018 and 2019, supplemented with additional tweets from Archive.org which cover the same period but are not evenly distributed.


\noindent \textbf{Data cleaning.}
Before training any model, we filter each model's training set of tweets using the procedure detailed in this section. Starting with the assumption that bots are amongst the most active users, we remove tweets from the top one percent of users that have posted most frequently. 
Additionally, following the recommendation of \citet{lee2021deduplicating}, we remove duplicates and near-duplicates. We find near-duplicates by hashing the texts of tweets after lowercasing and stripping punctuation. Hashing is performed using MinHash \citep{666900}, with 16 permutations.
Finally, user mentions are replaced with a generic placeholder (@user), except for verified users.

\subsection{Language model training}

Once the Twitter corpus has been collected and cleaned, we proceed to the language model pre-training. This consists of two phases: (1) training of a base model consisting of data until the end of 2019; 
and (2) continual training of language models every three months since the date of the base model. 


\noindent \textbf{Base model training.}
Our base model is trained with data until 2019 (included). Following \newcite{barbieri-etal-2020-tweeteval}, we start from the original RoBERTa-base model \cite{liu2019roberta} and continue training the masked language model on Twitter data. The model is trained using the same settings as \newcite{barbieri-etal-2020-tweeteval}, namely early stopping on the validation split and a learning rate of $1.0e^{-5}$. This initial 2019-90M base model converged after around fifteen days on 8 NVIDIA V100 GPUs.


\noindent \textbf{Continuous training.}
After training our base model, our goal is to continue training this language model with recent Twitter corpora. At the time of writing, for practical and logistical reasons, the decision is to train a new version of each language model every three months. The process to train this updated language model is simple, as it follows the same training procedure as the initial pre-training of the language model explained above. Our commitment is to keep updating and releasing a new model every three months, effectively enabling the community to make use of an up-to-date language model at any period in time.


\subsection{TimeLMs release summary}
\label{summary}

In Table \ref{tab:corpusstats} we include a summary of the Twitter corpora collected and models trained until the date of writing. Models are split in four three-month quarters (Q1, Q2, Q3 and Q4). Our base 2019-90M model consists of 90 million tweets until the end of 2019. Then, every quarter (i.e., every three months) 4.2M additional tweets are added, and the model gets updated as described above. Our latest released models, which are 2021-Q4 and 2021-124M (the latter was re-trained only once with all the data from 2020 and 2021), are trained on 124M tweets on top of the original RoBERTa-base model \cite{liu2019roberta}. All models are currently available through the Hugging Face hub at \url{https://huggingface.co/cardiffnlp}.

\begin{table}[ht]
\centering
\begin{tabular}{lcc}
\toprule
\textbf{Models}    & \textbf{Additional} & \textbf{Total}   \\ \midrule \midrule
2019-90M  & -          & 90.26M  \\ \midrule
2020-Q1   & 4.20M      & 94.46M  \\
2020-Q2   & 4.20M      & 98.66M  \\
2020-Q3   & 4.20M      & 102.86M \\
2020-Q4   & 4.20M      & 107.06M \\
2021-Q1   & 4.20M      & 111.26M \\
2021-Q2   & 4.20M      & 115.46M \\
2021-Q3   & 4.20M      & 119.66M \\
2021-Q4   & 4.20M      & 123.86M \\ \midrule
2021-124M & 33.60M      & 123.86M \\
\bottomrule
\end{tabular}
\caption{Number of tweets used to train each model. Showing number of tweets used to update models, and total starting from RoBERTa-base by \citet{liu2019roberta}.}
\label{tab:corpusstats}
\end{table}

In addition to these corpora for training language models, we set apart a number of tweets for each quarter (independent from the training set, with no overlap). These sets are used as test sets on our perplexity evaluation (see Section \ref{pereval}), and consist of 300K tweets per quarter, which were sampled and cleaned in the same way as the original corpus. 

\section{Evaluation}

In this section, we aim at evaluating the effectiveness of time-specific language models (see Section \ref{sec:timelms}) on time-specific tasks. In other words, our goal is to test the possible degradation of older models over time and, accordingly, test if this can be mitigated by continuous training.


\noindent \textbf{Evaluation tasks.}
We evaluated the released language models in two tasks: (1) TweetEval \cite{barbieri-etal-2020-tweeteval}, which consists of seven downstream tweet classification tasks; and (2) Pseudo-perplexity on corpora sampled from different time periods. While the first evaluation is merely aimed at validating the training procedure of the base language model, the second evaluation is the core contribution of this paper in terms of evaluation, where different models can be tested in different time periods.  

\subsection{TweetEval}

TweetEval \cite{barbieri-etal-2020-tweeteval} is a unified Twitter benchmark composed of seven heterogeneous tweet classification tasks. It is commonly used to evaluate the performance of language models (or task-agnostic models more generally) on Twitter data. With this evaluation, our goal is simply to show the general competitiveness of the models released with our package, irrespective of their time periods.

\noindent \textbf{Evaluation tasks.} The seven tweet classification tasks in TweetEval are emoji prediction \cite{barbieri-etal-2018-semeval}, emotion recognition \cite{mohammad-etal-2018-semeval}, hate speech detection \cite{basile-etal-2019-semeval}, irony detection \cite{van-hee-etal-2018-semeval}, offensive language identification \cite{zampieri-etal-2019-semeval}, sentiment analysis \cite{rosenthal-etal-2017-semeval} and stance detection \cite{mohammad-etal-2016-semeval}.

\noindent \textbf{Experimental setting.} Similarly to the TweetEval original baselines, only a moderate parameter search was conducted. The only hyper-parameter fine-tuned was the learning rate ($1.0e^{-3}$, $1.0e^{-4}$, $1.0e^{-5}$). The number of epochs each model is trained is variable, as we used early stopping monitoring the validation loss. The validation loss is also used to select the best model in each task.

\noindent \textbf{Comparison systems.} The comparison systems (SVM, FastText, BLSTM, RoBERTa-base and TweetEval) are those taken from the original TweetEval paper, as well as the state-of-the-art BERTweet model \cite{nguyen-etal-2020-bertweet}, which is trained over 900M tweets (posted between 2013 and 2019). All the language models compared are based on the RoBERTa-base architecture. 

\begin{table*}[!t] 
\centering
\renewcommand{\arraystretch}{1}
\setlength{\tabcolsep}{2.0pt}
\resizebox{0.87\textwidth}{!}{ 

\begin{tabular}{c|c|c|c|c|c|c|c||c}
\toprule
  \textbf{} &
  \textbf{Emoji} &
  \textbf{Emotion} &
  \textbf{Hate} &
  \textbf{Irony} &
  \textbf{Offensive} &
  \textbf{Sentiment} &
  \textbf{Stance} &
  \textbf{ALL} \\ \hline \hline

  SVM &
  29.3 &
  64.7 &
  36.7 &
  61.7 &
  52.3 &
  62.9 &
  67.3 &
  53.5 \\
  FastText &
  25.8 &
  65.2 &
  50.6 &
  63.1 &
  73.4 &
  62.9 &
  65.4 &
 58.1 \\

  BLSTM &
  24.7 &
  66.0 &
  52.6 &
  62.8 &
  71.7 &
  58.3 &
  59.4 &
  56.5  \\
  RoBERTa-Base &
  30.8 &
  76.6 &
  44.9 &
  55.2 &
  78.7 &
  72.0 &
  70.9 &
  61.3 \\
  TweetEval &
  31.6 &
  79.8 &
  55.5 &
  62.5 &
  81.6 &
  72.9 &
  72.6 &
  65.2 \\ 
  BERTweet &
  33.4 &
  79.3 &
  56.4 &
  \textbf{82.1} &
  79.5 &
  73.4 &
  71.2 &
  67.9 \\ 
  TimeLM-19 &
  33.4 &
  \textbf{81.0} &
  \textbf{58.1} &
  48.0 &
  \textbf{82.4} &
  73.2 &
  70.7 &
  63.8 \\
  TimeLM-21 &
  \textbf{34.0} &
  80.2 &
  55.1 &
  64.5 &
  82.2 &
  \textbf{73.7} &
  \textbf{72.9} &
  66.2 \\
  \hline \hline
\multicolumn{1}{c|}{\textbf{Metric}} &
  M-F1 &
  M-F1 &
  M-F1 &
  F$^{(i)}$ &
  M-F1 &
  M-Rec &
  AVG (F1) & TE
   \\ \bottomrule
\end{tabular}
}
\caption{\label{table-results-tweeteval} TweetEval test results of all comparison systems. 
}
\end{table*}

\noindent \textbf{Results.} TweetEval results are summarized in Table \ref{table-results-tweeteval}. BERTweet, which was trained on substantially more data, attains the best averaged results. However, when looking at single tasks, BERTweet outperforms both our latest released models, i.e., TimeLM-19 and TimeLM-21, on the irony detection task\footnote{We note that the irony dataset was created via distant supervision using the \#irony hashtag, and there could be a ``labels'' leak since BERTweet was the only model trained on tweets of the time period (2014/15) of the irony dataset.} only. It is also important to highlight that TweetEval tasks include tweets dated until 2018 at the latest (with most tasks being considerably earlier). This suggests that our latest released model (i.e. TimeLM-21), even if trained up to 2021 tweets, is generally competitive even on past tweets. Indeed, TimeLM-21 outperforms the most similar TweetEval model, which was trained following a similar strategy (in this case trained on fewer tweets until 2019), in most tasks.

\subsection{Time-aware language model evaluation}
\label{pereval}

Once the effectiveness of the base and subsequent models have been tested in downstream tasks, our goal is to measure to what extent the various models released are sensitive to a more time-aware evaluation. To this end, we rely on the pseudo perplexity measure \cite{salazar-etal-2020-masked}.

\noindent \textbf{Evaluation metric: Pseudo-perplexity (PPPL).}
The pseudo log-likelihood (PLL) score introduced by \citet{salazar-etal-2020-masked} is computed by iteratively replacing each token in a sequence with a mask, and summing the corresponding conditional log probabilities. This approach is specially suited to masked language models, rather than traditional left-to-right models. Pseudo-perplexity (PPPL) follows analogously from the standard perplexity formula, using PLL for conditional probability.

\noindent \textbf{Results.}
Table \ref{tab:pseudo_ppls} shows the pseudo-perplexity results in all test sets. As the main conclusion, the table shows how more recent models tend to outperform models trained when evaluated older data in most test sets (especially those contemporaneous). This can be appreciated by simply observing the decreasing values in the columns of the Table \ref{tab:pseudo_ppls}. There are a few interesting exceptions, however. For instance, the 2020-Q1 and 2020-Q2 test sets, which corresponding to the global start of the coronavirus pandemic, are generally better suited for models trained until that periods. Nonetheless, models trained on more contemporary data appear to converge to the optimal results. 

\begin{table*}[ht]
\centering
\resizebox{\textwidth}{!}{%
\begin{tabular}{l|cccccccc|c}
\toprule
\textbf{Models}  & \textbf{2020-Q1}                    & \textbf{2020-Q2}                    & \textbf{2020-Q3}                    & \textbf{2020-Q4}                    & \textbf{2021-Q1}                    & \textbf{2021-Q2}                    & \textbf{2021-Q3}                             & \textbf{2021-Q4}                             & \textbf{Change}           \\ \midrule \midrule
\textbf{\citealp{barbieri-etal-2020-tweeteval}} & \cellcolor[HTML]{FFFFFF}9.420       & \cellcolor[HTML]{FFFFFF}9.602       & \cellcolor[HTML]{FFFFFF}9.631       & \cellcolor[HTML]{FFFFFF}9.651       & \cellcolor[HTML]{FFFFFF}9.832       & \cellcolor[HTML]{FFFFFF}9.924       & \cellcolor[HTML]{FFFFFF}10.073               & \cellcolor[HTML]{FFFFFF}10.247               & N/A           \\ \midrule
\textbf{2019-90M}    & 4.823                               & 4.936                               & 4.936                               & 4.928                               & 5.093                               & 5.179                               & 5.273                                        & 5.362                                        &  N/A            \\ \midrule
\textbf{2020-Q1} & \cellcolor[HTML]{E8F5EF}4.521       & \cellcolor[HTML]{FFFFFF}4.625       & \cellcolor[HTML]{FFFFFF}4.699       & \cellcolor[HTML]{FFFFFF}4.692       & \cellcolor[HTML]{FFFFFF}4.862       & \cellcolor[HTML]{FFFFFF}4.952       & \cellcolor[HTML]{FFFFFF}5.043                & \cellcolor[HTML]{FFFFFF}5.140                & {-} \\
\textbf{2020-Q2} & \cellcolor[HTML]{57BB8A}{\ul 4.441} & \cellcolor[HTML]{57BB8A}{\ul 4.439} & \cellcolor[HTML]{BAE3CF}4.548       & \cellcolor[HTML]{C8E9D9}4.554       & \cellcolor[HTML]{CFEBDD}4.716       & \cellcolor[HTML]{D2EDE0}4.801       & \cellcolor[HTML]{DAF0E5}4.902                & \cellcolor[HTML]{DDF1E7}5.005                & -4.01\%                        \\
\textbf{2020-Q3} & \cellcolor[HTML]{FFFFFF}4.534       & \cellcolor[HTML]{A4DABF}4.525       & \cellcolor[HTML]{8DD1B0}4.450       & \cellcolor[HTML]{AEDEC6}4.487       & \cellcolor[HTML]{BAE3CF}4.652       & \cellcolor[HTML]{C0E5D3}4.738       & \cellcolor[HTML]{C7E8D8}4.831                & \cellcolor[HTML]{CEEBDD}4.945                & -2.15\%                        \\
\textbf{2020-Q4} & \cellcolor[HTML]{FDFEFD}4.533       & \cellcolor[HTML]{A3D9BF}4.524       & \cellcolor[HTML]{84CDA9}4.429       & \cellcolor[HTML]{7DCAA4}4.361       & \cellcolor[HTML]{9FD8BC}4.571       & \cellcolor[HTML]{ACDDC5}4.672       & \cellcolor[HTML]{B5E1CB}4.763                & \cellcolor[HTML]{B9E2CE}4.859                & -2.81\%                        \\
\textbf{2021-Q1} & \cellcolor[HTML]{D2ECDF}4.509       & \cellcolor[HTML]{8DD1AF}4.499       & \cellcolor[HTML]{76C79F}4.399       & \cellcolor[HTML]{72C69D}4.334       & \cellcolor[HTML]{73C69E}4.439       & \cellcolor[HTML]{8FD1B1}4.574       & \cellcolor[HTML]{9CD7BA}4.668                & \cellcolor[HTML]{A2D9BE}4.767                & -2.89\%                        \\
\textbf{2021-Q2} & \cellcolor[HTML]{BFE5D2}4.499       & \cellcolor[HTML]{7DCAA4}4.481       & \cellcolor[HTML]{6BC398}4.376       & \cellcolor[HTML]{6CC399}4.319       & \cellcolor[HTML]{6AC297}4.411       & \cellcolor[HTML]{69C297}4.445       & \cellcolor[HTML]{83CCA8}4.570                & \cellcolor[HTML]{8BD0AE}4.675                & -2.83\%                        \\
\textbf{2021-Q3} & \cellcolor[HTML]{8DD0AF}4.471       & \cellcolor[HTML]{64C093}4.455       & \cellcolor[HTML]{59BB8B}4.335       & \cellcolor[HTML]{5DBD8E}4.280       & \cellcolor[HTML]{5BBD8D}4.366       & \cellcolor[HTML]{5ABC8C}4.394       & \cellcolor[HTML]{5CBD8D}4.422                & \cellcolor[HTML]{70C59B}4.565                & -3.26\%                        \\
\textbf{2021-Q4} & \cellcolor[HTML]{86CEAB}4.467       & \cellcolor[HTML]{65C093}4.455       & \cellcolor[HTML]{57BB8A}{\ul 4.330} & \cellcolor[HTML]{57BB8A}{\ul 4.263} & \cellcolor[HTML]{57BB8A}{\ul 4.351} & \cellcolor[HTML]{57BB8A}{\ul 4.381} & \cellcolor[HTML]{57BB8A}{\ul \textbf{4.402}} & \cellcolor[HTML]{57BB8A}{\ul \textbf{4.463}} & -2.24\%                        \\ \midrule
\textbf{2021-124M} & \textbf{4.319}                      & \textbf{4.297}                      & \textbf{4.279}                      & \textbf{4.219}                      & \textbf{4.322}                      & \textbf{4.361}                      & 4.404                                        & 4.489                                        & N/A          \\
\bottomrule
\end{tabular}%
}
\caption{Pseudo-perplexity results (lower is better) of all models in the Twitter test sets sampled from different quarters (each quarter correspond to three months. Q1: Jan-Mar; Q2: Apr-Jun; Q3: Jul-Sep; Q4: Oct-Dec). The last column reports difference in pseudo-perplexity, comparing the value obtained for each quarter's test set, between the model trained on the previous quarter and the model updated with data from that same quarter.}
\label{tab:pseudo_ppls}
\end{table*}

\noindent \textbf{Degradation over time.} How long does it take for a model to be outdated? 
Overall, PPPL scores tend to increase almost 10\% after one year. In general, PPPL appears to decrease consistently every quarterly update. 
This result reinforces the need for updated language models even for short time periods such as three-month quarters. In most cases, degradation on future data is usually larger than on older data. This result is not completely unexpected since newer models are also trained on more data for more time periods. In Section \ref{degradation} we expand on this by including a table detailing the relative performance degradation over language models over time. 

\vspace{-0.1cm}

\section{Python Interface}
\vspace{-0.2cm}

In this section we present an integrated Python interface that we release along with the data and language models presented in this paper. As mentioned in Section \ref{summary}, all language models will be available from the Hugging Face hub and our code is designed to be used with this platform. 

Our interface, based on the Transformers package \cite{wolf-etal-2020-transformers}, is focused on providing easy single-line access to language models trained for specific periods and related use cases.
The choice of language models to be used with our interface is determined using one of four modes of operation:
(1) `latest': using our most recently trained Twitter model; (2) `corresponding': using the model that was trained only until each tweet's date (i.e., its specific quarter); (3) custom: providing the preferred date or quarter (e.g., `2021-Q3'); and (4) `quarterly': using all available models trained over time in quarterly intervals.
Having specified the preferred language models, there are three main functionalities within the code, namely: (1) computing pseudo-perplexity scores, (2) evaluating language models in our released or customized test sets, and (3) obtaining masked predictions.

Users can measure the extent to which the chosen pretrained language models are aligned (i.e., familiar) with a given list of tweets (or any text) using pseudo-perplexity (see Section \ref{pereval} for more details), computed as shown in Code \ref{codeppl}.




\begin{lstlisting}[numbers=none, language=Python, label=codeppl, caption=Computing Pseudo-PPL on a given tweet using the most recently available model.]
from timelms import TimeLMs
tlms = TimeLMs(device='cuda')

tweets = [{'text': 'Looking forward to watching Squid Game tonight !'}]

pseudo_ppls = tlms.get_pseudo_ppl(tweets,
    mode='latest') # loads 2021-Q4 model
\end{lstlisting}

For a more extensive evaluation of language models using pseudo-perplexity, we provide a random subset of our test data across 2020 and 2021.\footnote{Limited to 50K tweets, the maximum allowed by Twitter. IDs for all test tweets are available on the repository.}
To evaluate other models from the Transformers package, we provide the `eval\_model' method (\texttt{tlms.eval\_model()}) to compute pseudo-perplexity on any given set of tweets or texts (e.g., the subset we provide) using other language models supported by the Transformers package.
Both scoring methods not only provide the pseudo-perplexity scores specific to each model (depending on specified model name, or TimeLMs specified mode), but also the PLL scores assigned to each tweet by the different models.






Finally, predictions for masked tokens of any given tweet or text may be easily obtained as demonstrated in Code \ref{code1}. 

\begin{lstlisting}[numbers=none, language=Python, label=code1, caption=Obtaining masked predictions using model corresponding to the tweet's date. Requires tweets or texts with a \texttt{<mask>} token.]
tweets = [{"text": "So glad I'm <mask> vaccinated.", "created_at": "2021-02-01T23:14:26.000Z"}]

preds = tlms.get_masked_predictions(tweets, top_k=3,
    mode='corresponding')  # loads 2021-Q1 model
\end{lstlisting}

Note that while the examples included in this paper are associated with specific dates (i.e., the \texttt{created\_at}  field), these are only required for the `corresponding' mode.

\section{Analysis}

To complement the evaluation in the previous section, we perform a more detailed analysis in three important aspects: (1) a quantitative analysis on the degradation suffered by language models over time; (2) the relation between time and size (Section \ref{controlsize}); and (3) a qualitative analysis where we show the influence of time in language models for specific examples (Section \ref{qualitative}). 

\subsection{Degradation analysis}
\label{degradation}

Table \ref{tab:pppl_variation} displays the relative performance degradation (or improvement) of TimeLMs language models with respect to the test sets whose time period is the latest where they have been trained on (diagonals in the table). The table shows how models tend to perform worse in newer data sets, with a degradation of performance up to 13.68\% of the earlier 2020-Q1 model on the latest 2021-Q4 model (with data almost two years later than the latest data the language model was trained on).

\begin{figure*}[tb]
\centering
\resizebox{0.8\textwidth}{!}{
\includegraphics[]{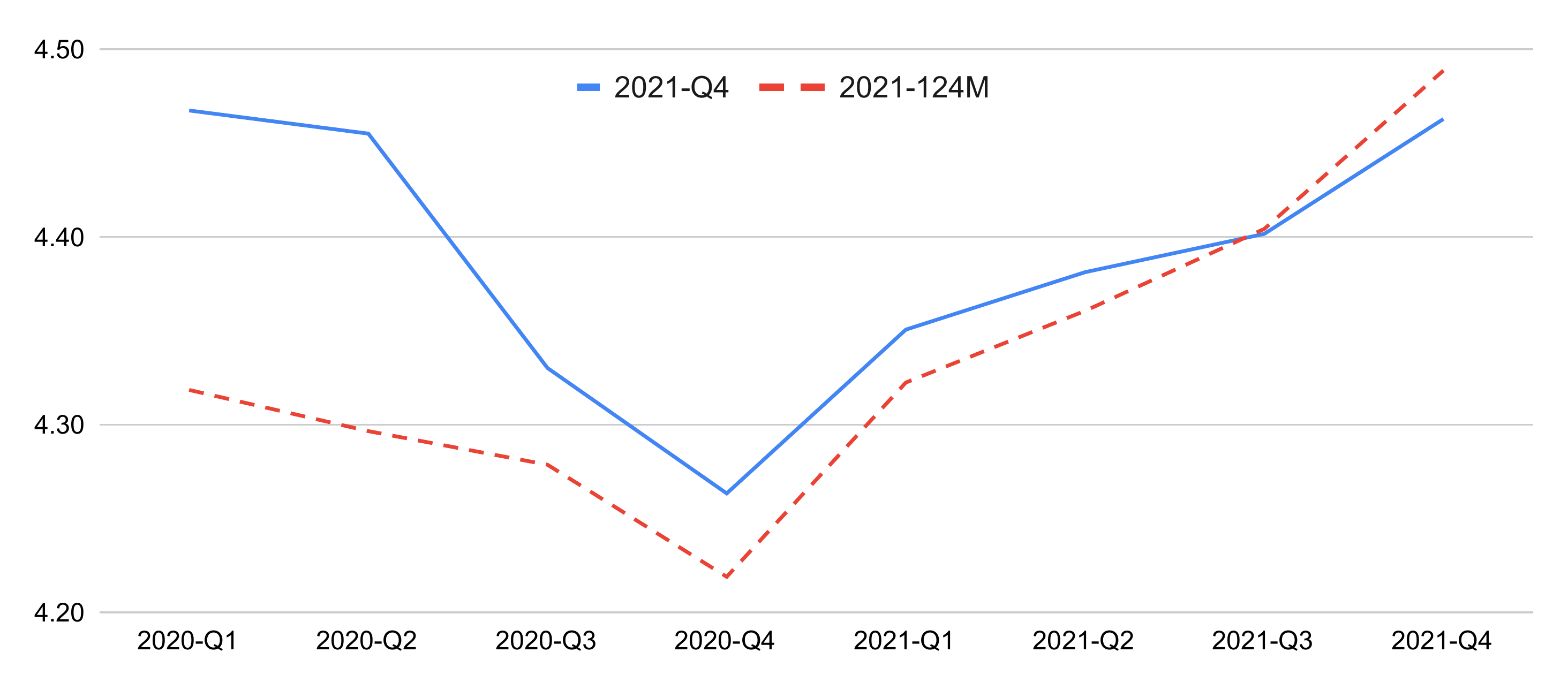}
}
\caption{Performance (PPPL) of 2021-124M and 2021-Q4 models across the test sets.}
\label{fig:chart}
\end{figure*}

In order to compare the effect of continuous training with respect to single training, Figure \ref{fig:chart} shows the PPPL performances of 2021-124M (trained on all 2020-2021 data at once) and the 2021-Q4 (updating 2021-Q3) models. Note how 2021-124M shows improved performance generally, with the largest differences being attained on the first two quarters of 2020, but not for the latest quarters where continuous training seems to work slightly better. While more analysis would be required, this result suggests that a single training is beneficial for earlier periods, while a quarterly training seems to be better adapted to the most recent data. However, there does not seem to be any meaningful catastrophic forgetting in the quarterly-updated model, as the differences are relative small.

\begin{table*}[tb]
\centering
\resizebox{0.80\textwidth}{!}{%
\begin{tabular}{l|cccccccc}
\toprule
\textbf{Models}  & \textbf{2020-Q1}               & \textbf{2020-Q2}                & \textbf{2020-Q3}                & \textbf{2020-Q4}                & \textbf{2021-Q1}                & \textbf{2021-Q2}                & \textbf{2021-Q3}                & \textbf{2021-Q4}                \\ \midrule \midrule
\textbf{2020-Q1} & \cellcolor[HTML]{FFFFFF}0.00\% & \cellcolor[HTML]{FBEAE8}2.29\%  & \cellcolor[HTML]{F8DAD7}3.94\%  & \cellcolor[HTML]{F9DBD9}3.78\%  & \cellcolor[HTML]{F2B7B2}7.52\%  & \cellcolor[HTML]{EEA49E}9.52\%  & \cellcolor[HTML]{EA9189}11.53\% & \cellcolor[HTML]{E67C73}13.68\% \\
\textbf{2020-Q2} & \cellcolor[HTML]{FFFFFF}0.04\% & \cellcolor[HTML]{FFFFFF}0.00\%  & \cellcolor[HTML]{FBE6E4}2.46\%  & \cellcolor[HTML]{FAE5E3}2.59\%  & \cellcolor[HTML]{F3BFBB}6.24\%  & \cellcolor[HTML]{F0ACA6}8.16\%  & \cellcolor[HTML]{EB948D}10.42\% & \cellcolor[HTML]{E67C73}12.75\% \\
\textbf{2020-Q3} & \cellcolor[HTML]{FBE9E8}1.87\% & \cellcolor[HTML]{FCECEA}1.67\%  & \cellcolor[HTML]{FFFFFF}0.00\%  & \cellcolor[HTML]{FEF6F5}0.82\%  & \cellcolor[HTML]{F5CAC6}4.53\%  & \cellcolor[HTML]{F1B3AE}6.47\%  & \cellcolor[HTML]{EC9B94}8.54\%  & \cellcolor[HTML]{E67C73}11.10\% \\
\textbf{2020-Q4} & \cellcolor[HTML]{F7D2CF}3.95\% & \cellcolor[HTML]{F7D5D2}3.74\%  & \cellcolor[HTML]{FCEEEC}1.57\%  & \cellcolor[HTML]{FFFFFF}0.00\%  & \cellcolor[HTML]{F5C8C4}4.82\%  & \cellcolor[HTML]{F0AEA8}7.14\%  & \cellcolor[HTML]{EB968F}9.22\%  & \cellcolor[HTML]{E67C73}11.43\% \\
\textbf{2021-Q1} & \cellcolor[HTML]{FAE3E1}1.58\% & \cellcolor[HTML]{FBE7E6}1.37\%  & \cellcolor[HTML]{BFE5D3}-0.89\% & \cellcolor[HTML]{57BB8A}-2.36\% & \cellcolor[HTML]{FFFFFF}0.00\%  & \cellcolor[HTML]{F5C9C6}3.05\%  & \cellcolor[HTML]{EEA49E}5.16\%  & \cellcolor[HTML]{E67C73}7.39\%  \\
\textbf{2021-Q2} & \cellcolor[HTML]{FAE1DF}1.21\% & \cellcolor[HTML]{FCEBE9}0.82\%  & \cellcolor[HTML]{A3D9BF}-1.55\% & \cellcolor[HTML]{57BB8A}-2.83\% & \cellcolor[HTML]{D1ECDF}-0.77\% & \cellcolor[HTML]{FFFFFF}0.00\%  & \cellcolor[HTML]{F2B8B3}2.83\%  & \cellcolor[HTML]{E67C73}5.19\%  \\
\textbf{2021-Q3} & \cellcolor[HTML]{F7D2CF}1.12\% & \cellcolor[HTML]{FAE1DF}0.75\%  & \cellcolor[HTML]{98D5B7}-1.95\% & \cellcolor[HTML]{57BB8A}-3.20\% & \cellcolor[HTML]{BCE4D0}-1.26\% & \cellcolor[HTML]{DEF1E8}-0.61\% & \cellcolor[HTML]{FFFFFF}0.00\%  & \cellcolor[HTML]{E67C73}3.25\%  \\
\textbf{2021-Q4} & \cellcolor[HTML]{E67C73}0.10\% & \cellcolor[HTML]{F8FCFA}-0.17\% & \cellcolor[HTML]{8FD1B1}-2.97\% & \cellcolor[HTML]{57BB8A}-4.47\% & \cellcolor[HTML]{A0D8BD}-2.51\% & \cellcolor[HTML]{BAE3CF}-1.83\% & \cellcolor[HTML]{CBEADB}-1.37\% & \cellcolor[HTML]{FFFFFF}0.00\% \\
\bottomrule
\end{tabular}%
}
\caption{Difference across quarterly models and test sets comparing the pseudo-perplexity observed at the quarter corresponding to each model, against the pseudo-perplexity observed for that same model on both previous and future test sets. Highlights model degradation on future data, as well as how models fare on past data.}
\label{tab:pppl_variation}
\end{table*}

\subsection{Time and size control experiment}
\label{controlsize}

Given the results presented earlier, one may naturally wonder whether the improvement may be due to the increase in training size or the recency of additional data. While this question is not easy to answer (and probably the answer will be in-between these two reasons), we perform a simple control experiment as an initial attempt. 
To this end, we trained an additional language model with twice the training data of the third quarter of 2021 (2021-Q3). This way, the total number of training tweets is exactly the same as the model trained until the fourth quarter of 2021 (2021-Q4).

\begin{table}[t]
\centering
\resizebox{0.9\columnwidth}{!}{%
\begin{tabular}{l|ccc}
\toprule
\textbf{Models}      & \textbf{2021-Q2}     & \textbf{2021-Q3}     & \textbf{2021-Q4}     \\ \midrule \midrule
\textbf{2021-Q2}     & 4.445                & 4.570                & 4.675                \\ \midrule
\textbf{2021-Q3}     & 4.394                & 4.422                & 4.565                \\
\textbf{2021-Q3-2x}  & {\ul \textbf{4.380}} & {\ul \textbf{4.380}} & 4.534                \\ \midrule
\textbf{2021-Q4}     & 4.381                & 4.402                & {\ul \textbf{4.463}} \\
\bottomrule
\end{tabular} %
}
\caption{Results of the control experiment comparing quarterly models where the 2021-Q3 model is trained with twice the data from that quarter (2021-Q3-2x).}
\label{tab:pppl_extra}
\vspace{-0.07cm}
\end{table}

Considering the results on Table \ref{tab:pppl_extra}, we find that the model trained on twice the data for Q3 outperforms the model trained with the default Q3 data in all tested quarters. This confirms the assumption that increasing training data leads to improved language model performance. When comparing with the model trained until 2021-Q4, results show this 2021-Q3-2x model is only slightly better in the 2021-Q2 and 2021-Q3 test sets. However, as we could expect, the model trained in more recent data (i.e., until 2021-Q4) gets the best overall results on the more recent test set (i.e., 2021-Q4). 


\subsection{Qualitative analysis}
\label{qualitative}

\begin{table}[t]
\centering
\resizebox{1.0\columnwidth}{!}{
\begin{tabular}{@{}c|c|c|c@{}}
\toprule
\textbf{Model} &
    \textbf{\shortstack{So glad \\ I'm  \textless{}mask\textgreater \\ vaccinated.}} &
  \textbf{\shortstack{I keep \\ forgetting to \\ bring a \textless{}mask\textgreater .}} &
  \textbf{\shortstack{Looking forward \\ to watching  \textless{}mask\textgreater \\ Game tonight!}} \\ \cmidrule(l){1-4} 
\multicolumn{1}{c|}{\multirow{3}{*}{\textbf{2020-Q1}}}        & not     & bag     & the   \\
\multicolumn{1}{c|}{}                                     & getting & purse   & The   \\
\multicolumn{1}{c|}{}                                     & self    & charger & this  \\ \midrule
\multicolumn{1}{c|}{\multirow{3}{*}{\textbf{2020-Q2}}}        & not     & mask    & The   \\
\multicolumn{1}{c|}{}                                     & getting & bag     & the   \\
\multicolumn{1}{c|}{}                                     & fully   & purse   & End   \\ \midrule
\multicolumn{1}{c|}{\multirow{3}{*}{\textbf{2020-Q3}}}        & not     & mask    & the   \\
\multicolumn{1}{c|}{}                                     & getting & bag     & The   \\
\multicolumn{1}{c|}{}                                     & fully   & purse   & End   \\ \midrule
\multicolumn{1}{c|}{\multirow{3}{*}{\textbf{2020-Q4}}} & not     & bag     & the   \\
\multicolumn{1}{c|}{}                                     & getting & purse   & The   \\
\multicolumn{1}{c|}{}                                     & fully   & charger & End   \\ \midrule
\multicolumn{1}{c|}{\multirow{3}{*}{\textbf{2021-Q1}}}        & getting & purse   & the   \\
\multicolumn{1}{c|}{}                                     & not     & charger & The   \\
\multicolumn{1}{c|}{}                                     & fully   & bag     & End   \\ \midrule
\multicolumn{1}{c|}{\multirow{3}{*}{\textbf{2021-Q2}}}        & fully   & bag     & the   \\
\multicolumn{1}{c|}{}                                     & getting & charger & The   \\
\multicolumn{1}{c|}{}                                     & not     & lighter & this  \\ \midrule
\multicolumn{1}{c|}{\multirow{3}{*}{\textbf{2021-Q3}}}        & fully   & charger & the   \\
\multicolumn{1}{c|}{}                                     & getting & bag     & The   \\
\multicolumn{1}{c|}{}                                     & not     & purse   & This  \\ \midrule
\multicolumn{1}{c|}{\multirow{3}{*}{\textbf{2021-Q4}}}            & fully   & bag     & Squid \\
\multicolumn{1}{c|}{}                                     & getting & lighter & the   \\
\multicolumn{1}{c|}{}                                     & not     & charger & The  \\
\bottomrule
\end{tabular}
}
\caption{Masked token prediction over time using three example tweets as input (using mode=`quarterly').
For each quarterly model, the table displays the top-3 predictions ranked by their prediction probability.
}
\label{tab:masklm-examples}
\end{table}

In this section we illustrate, in practice, how models trained on different quarters perceive specific tweets.
First, we use their masked language modeling head to predict a \texttt{<mask>} token in context.
Table \ref{tab:masklm-examples} shows three tweets and associated predictions from each of our quarterly models.
The model belonging to the most pertinent quarter exhibits background knowledge more aligned to the trends of that period.
In the two COVID-related examples, 
we observe increasing awareness of the general notion of being \textit{fully vaccinated} (as opposed to \textit{not vaccinated}, the top prediction from the 2020-Q1 model) in the former, and, in the latter, two instances where \textit{forgetting a mask} is more likely than forgetting other apparel less related to a particular period, such as \textit{a charger}, \textit{a lighter} or \textit{a purse}.
Finally, note how, in the last example, ``Looking forward to watching \texttt{<mask>} Game tonight!", it is only in 2021-Q4 that predictions change substantially, when the model has been exposed to reactions to the "Squid Game" show, overlapping in time with its global release.

\begin{figure}[!t]
\resizebox{\columnwidth}{!}{
\includegraphics[]{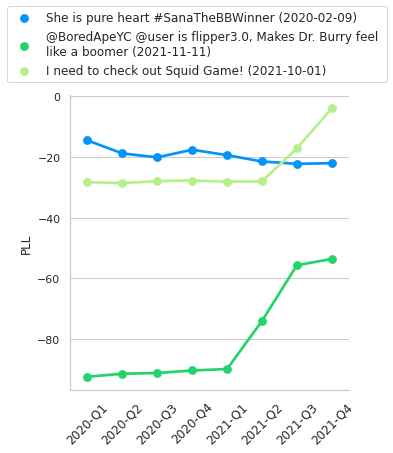}
}
\caption{PLL scores of TimeLMs language models trained over different periods for three selected tweets.}
\label{fig:pll}
\end{figure}

Our second piece of analysis involves the visualization of pseudo log-likehood (PLL) scores for tweets requiring awareness of a trend or event tied to a specific period (Figure \ref{fig:pll}). Indeed, more recent models are better at predicting tweets involving popular events, such as NFTs 
or, again, the show "Squid Game". Conversely, we observe a stagnation (or even degradation) of the PLL scores for a tweet about a contestant of an older reality show.

\section{Conclusion}

In this paper we presented TimeLMs, language models trained on Twitter over different time periods. The initiative also includes the future training of language models every three months, thus providing free-to-use and up-to-date language models for NLP practitioners. These language models are released together with a simple Python interface which facilitates loading and working with these models, including time-aware evaluation. In our evaluation in this paper, we have shown how time-aware training is relevant, not only from the theoretical point of view, but also the practical one, as the results demonstrate a clear degradation in performance when models are used for future data, which is one of the most common settings in practice.

As future work, we are planning to explicitly integrate the time span variable in the language models, i.e., introducing string prefixes, along the lines of \newcite{10.1162/tacl_a_00459} and \newcite{10.1145/3488560.3498529}.


\bibliography{anthology,custom}
\bibliographystyle{acl_natbib}



\end{document}